\documentclass[11pt]{article}

\usepackage[final]{acl}

\usepackage{times}
\usepackage{latexsym}
\usepackage[T1]{fontenc}
\usepackage[utf8]{inputenc}
\usepackage{microtype}
\usepackage{inconsolata}
\usepackage{graphicx}
\usepackage{booktabs}
\usepackage{subcaption}
\usepackage{amsmath}
\usepackage{amssymb}
\usepackage{xcolor}
\usepackage{tcolorbox}
\tcbuselibrary{skins,breakable}
\usepackage{placeins}
\usepackage{soul}

\definecolor{Acol}{HTML}{2A6FB4}
\definecolor{Bcol}{HTML}{B86541}
\definecolor{Abg}{HTML}{E6EEF7}
\definecolor{Bbg}{HTML}{F2D3C3}
\newcommand{\hlA}[1]{{\sethlcolor{Abg}\hl{#1}}}
\newcommand{\hlB}[1]{{\sethlcolor{Bbg}\hl{#1}}}
\newcommand{\seclbl}[1]{\textbf{\textsc{\footnotesize #1}}}
\newcommand{\rowlbl}[1]{\makebox[5em][l]{\textbf{\textsc{\footnotesize #1}}}}
\newcommand{\plbl}[1]{\makebox[4.5em][l]{\textbf{\textsc{\footnotesize #1}}}}

\newcommand{\plblA}[1]{\makebox[4.5em][l]{\textcolor{Acol}{\textbf{\textsc{\footnotesize #1}}}}}
\newcommand{\plblB}[1]{\makebox[4.5em][l]{\textcolor{Bcol}{\textbf{\textsc{\footnotesize #1}}}}}

\newtcolorbox{probebox}[1][]{
  enhanced, breakable,
  colback=gray!5, colframe=gray!50, boxrule=0.4pt,
  arc=2pt, left=4pt, right=4pt, top=3pt, bottom=3pt,
  fontupper=\small, #1
}

\newtcolorbox{splitbox}[1][]{
  enhanced, breakable,
  colback=gray!5, colbacklower=gray!5,
  colframe=gray!50, boxrule=0.4pt,
  arc=2pt, left=5pt, right=5pt, top=2.5pt, bottom=2.5pt,
  middle=1pt,
  fontupper=\small, fontlower=\small,
  segmentation style={solid, gray!55, line width=0.3pt},
  #1
}

\title{What Attention Recalls and Recurrence Controls \\
in Hybrid Language Models}

\author{
  Kirill Afendulev\textsuperscript{3,2}, \quad
  Alexey Dontsov\textsuperscript{3,1}, \quad
  Elena Tutubalina\textsuperscript{3,1}, \quad
  Anton Korznikov\textsuperscript{1,3} \\
  \textsuperscript{1}HSE University, Moscow, Russia \\
  \textsuperscript{2}YSDA, Moscow, Russia \\
  \textsuperscript{3}AIRI, Moscow, Russia \\[2pt]
  \textbf{Code:} \url{https://github.com/kirillTerra/split-prefill}
}

\begin{document}
\maketitle

\begin{abstract}
Hybrid language models combine attention with a fixed-size recurrent state, but the role of each channel remains unclear. We introduce two cache-level interventions. \emph{Split-prefill} keeps only the KV cache or only the recurrent state from a prefilled context, then generates an answer. \emph{State-swap} pairs the KV cache from one context with the recurrent state from another in a single forward pass. On Qwen3.5 and Falcon-H1, the two channels split sharply by function. Exact retrieval survives only through attention (64–98\% of full accuracy) and collapses to zero through recurrence. Output language and persona reverse the pattern: both survive recurrence (70–80\% and 3–5$\times$) while KV-only drops to ~1\% language accuracy. State-swap confirms this causally: the answer takes its value from the KV side and its language from the recurrent side. Recurrent-only generation also accepts words that were never in the context but share meaning or parts with seen items. Attention provides a lookup over what was said; the recurrent state shapes how the model says it next.
\end{abstract}

\section{Introduction}
\label{sec:intro}

State-space and linear-attention models are a practical alternative to full
self-attention for long-context language modelling
\citep{gu2023mamba,dao2024transformers,yang2023gated,katharopoulos2020transformers}.
Modern open-weight families such as Qwen3.5 \citep{yang2025qwen3},
Falcon-H1 \citep{zuo2025falcon}, and Jamba \citep{lieber2024jamba} combine
softmax attention with a fixed-size recurrent block such as Gated
DeltaNet \citep{yang2025gated} or Mamba \citep{gu2023mamba}. Once
such a model is trained, a mechanistic question follows: what does
each component actually carry at inference?

To test this we introduce two cache-level interventions
(\S\ref{sec:method}). \emph{Split-prefill} keeps either the KV cache
or the recurrent state from a context and drops the other;
\emph{state-swap} crosses the two caches from different contexts in
a single forward pass. We instantiate on Qwen3.5-4B (Gated DeltaNet)
and Falcon-H1-3B-Instruct (Mamba).

A natural first guess about the two channels is temporal: attention is
precise but expensive long-term memory, the recurrent state is a cheap
short-term buffer. This framing turns out to be misleading. Drop the
recurrent state and short-range processing still works, but the
response policy is gone: language, persona, instruction tone.
Dropping the attention KV cache has the opposite effect: response
policy is preserved, but the model can no longer report a value for
a specific key. The dissociation is functional: it concerns not the
age of information each channel retains, but the type of access each
channel provides at inference. The recurrent state collapses to
zero on exact retrieval, even though stand-alone state-space and
linear-attention models retain at least partial recall on the same
task families
\citep{arora2024zoology,jelassi2024repeat,park2024can}: in a trained
hybrid, the two components separate behaviourally under our
interventions.

\begin{figure*}[t]
  \centering
  \includegraphics[width=0.72\linewidth, trim=0 90 0 90, clip]{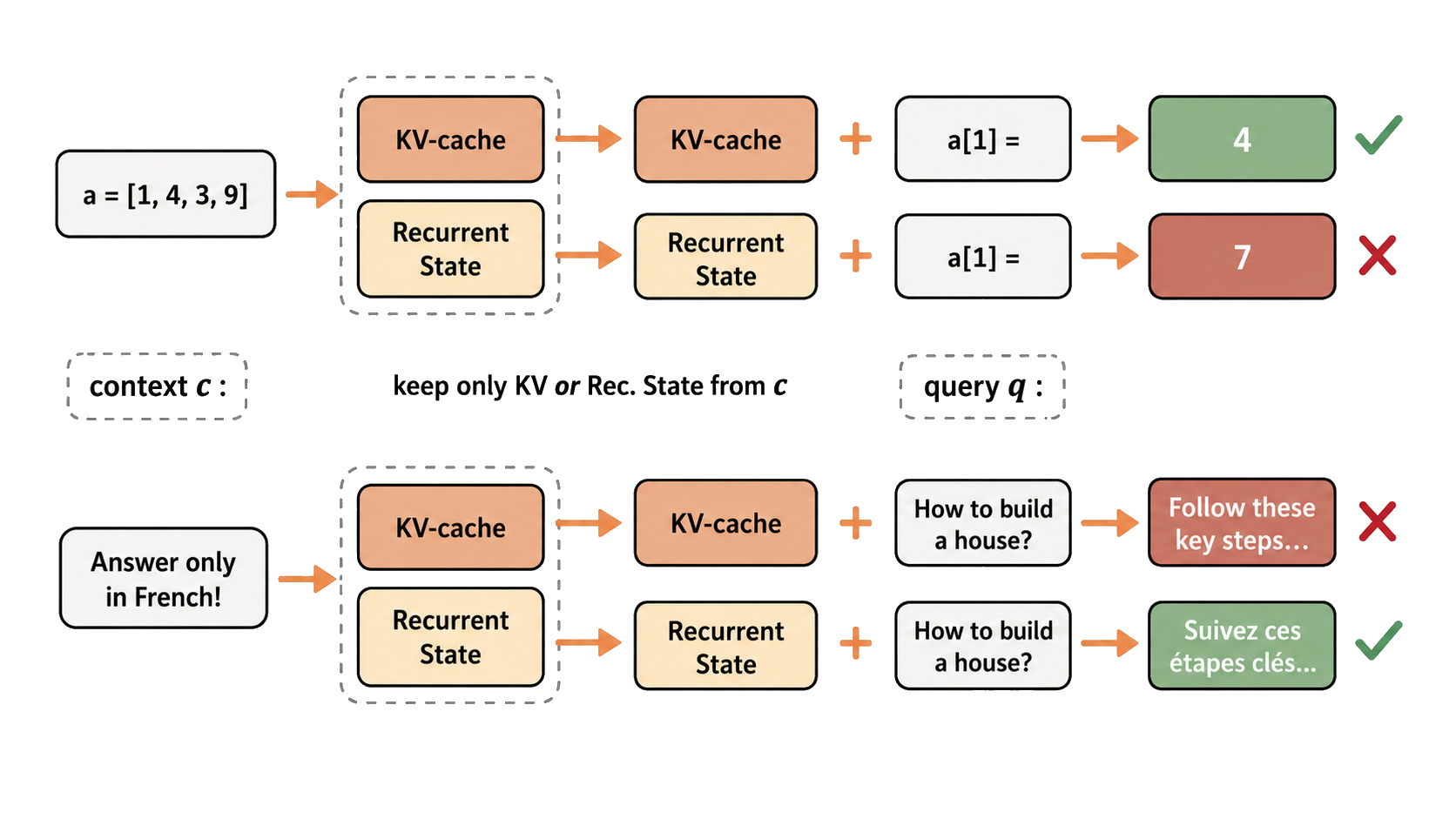}
    \vspace{-1em}

  \caption{\emph{Split-prefill}. A prompt is split into a context (\textsc{prompt
  1}) and a query (\textsc{prompt 2}). We prefill the context, save its two
  internal channels (attention KV cache, recurrent state), then continue on
  the query while keeping only one channel from the context and resetting
  the other.}
  \vspace{-1em}
  \label{fig:scheme}
\end{figure*}

\section{Related Work}
\label{sec:related}

\paragraph{Hybrid recurrent--attention LMs.}
State-space \citep{gu2023mamba,dao2024transformers}, linear-attention
\citep{katharopoulos2020transformers,schlag2021linear}, RWKV
\citep{peng2023rwkv}, GLA \citep{yang2023gated}, and Gated DeltaNet
\citep{yang2025gated} models all compress past context into a
fixed-size state. Strong recent open-weight models combine such a
block with standard attention in a hybrid stack
\citep{lieber2024jamba,yang2025qwen3,zuo2025falcon}: Jamba interleaves the
two block types, Qwen3.5 uses a $3{:}1$ Gated DeltaNet to Gated
Attention pattern, and Falcon-H1 adopts a parallel-hybrid block that
runs attention and SSMs alongside each other. Reported gains over
pure variant follow  \citet{waleffe2024empirical}.

\paragraph{Recall as a known weakness of compressed state.}
Fixed-state models struggle with exact copying, key--value retrieval, and
needle-in-a-haystack lookup compared to attention
\citep{arora2024zoology,jelassi2024repeat,park2024can,ben2025decimamba,liu2024lost,hsieh2024ruler,kamradt2023niah}.
Our results agree on  this direction; we make the stronger claim that, the recurrent state does not function as an addressable
memory for retrieval at inference. What it does carry is examined in
\S\ref{sec:demo-drm} and \S\ref{sec:demo-conjunction}.

\paragraph{Mechanistic analysis.}
Where information lives in transformers has been mapped via circuits and
activation patching
\citep{elhage2021mathematical,olsson2022context,wang2022interpretability,meng2022locating,geva2023dissecting};
\citet{sharma2024locating} extend this to Mamba. We complement this line of work
by causally swapping the two components of a hybrid on the same prompt.

\section{Method}
\label{sec:method}

\paragraph{Split-prefill.} A prompt is the concatenation of a context $c$
and a query $q$. A forward pass over $c$ populates the attention KV cache
$\mathcal{K}(c)$ and the recurrent state $\mathcal{R}(c)$ (Gated DeltaNet
for Qwen, Mamba for Falcon). We then process $q$ in three conditions:
\textsc{full} (keep both), \textsc{kv-only} (keep $\mathcal{K}$, drop
$\mathcal{R}$), and \textsc{rec-only} (keep $\mathcal{R}$, drop
$\mathcal{K}$). Because $q$ is always processed normally, the
intervention isolates the carry-over from $c$ in a specific channel.

\paragraph{State-swap.} We prefill two contexts $c_A, c_B$ of the same
format but with different content, then continue a shared query $q$ on
the hybrid cache $(\mathcal{K}(c_A), \mathcal{R}(c_B))$ and its mirror.
Properties of the answer that follow $c_A$ are attributed to attention;
those that follow $c_B$, to recurrence (Fig.~\ref{fig:stateswap},
\S\ref{sec:demo-swap}). Implementation details are in
Appendix~\ref{app:impl}.

\begin{figure*}[!t]
  \centering
  \includegraphics[width=0.98\linewidth]{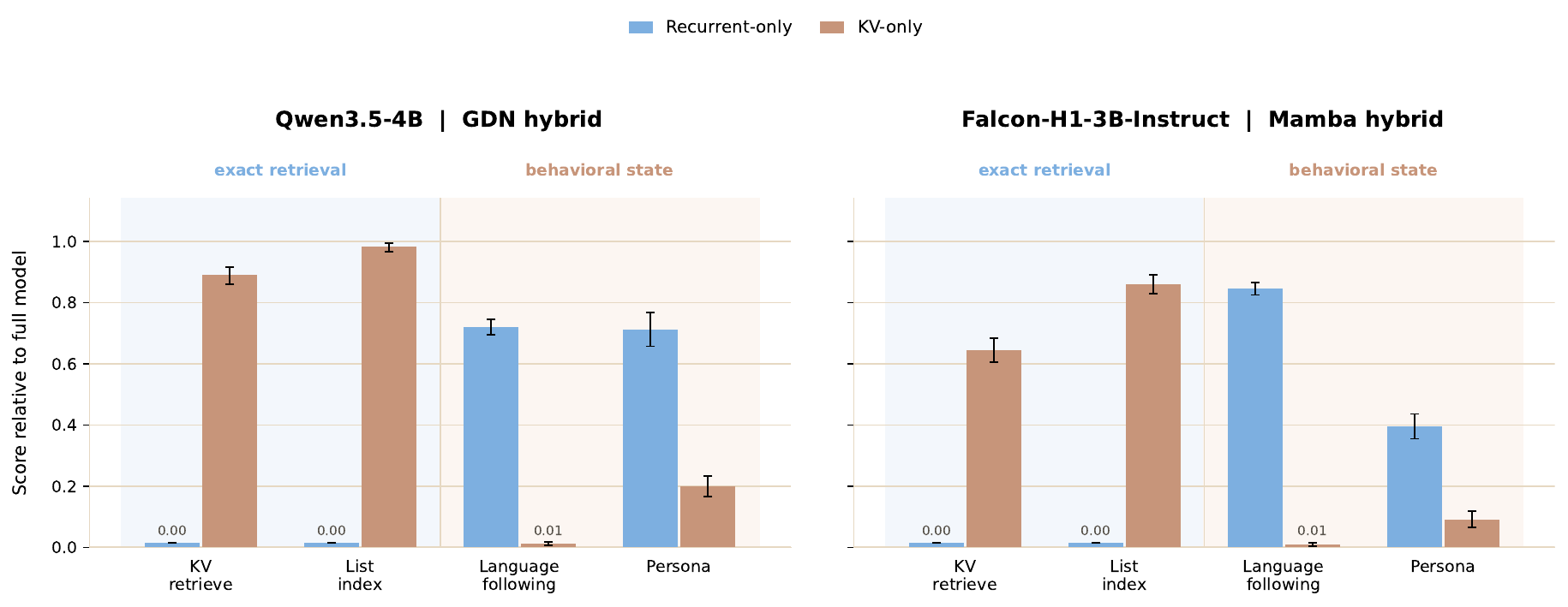}
  \caption{\textbf{Main result.} Split-prefill scores on four diagnostic
  tasks for Qwen3.5-4B (Gated DeltaNet hybrid, left) and
  Falcon-H1-3B-Instruct (Mamba hybrid, right), normalised to the
  full-model baseline. Token and position retrieval survive in the
  attention KV cache and collapse with recurrent state only. Language
  and persona reverse the pattern: they survive only the recurrent
  state and are nearly absent from KV alone. Numerical scores are in
  Table~\ref{tab:main}.}
  \label{fig:bars}
\end{figure*}

\section{Quantitative Dissociation}
\label{sec:quant}

\paragraph{Why small, isolated tasks.} Standard long-context benchmarks
demand token retrieval and response-mode maintenance at once, so
component ablations on them \citep{borobia2026functional} cannot
attribute any accuracy drop to a specific function. We instead use
four small synthetic tasks, each isolating one capability; the
trade-off is offset by the size of the effects we report, which are
large enough to support a causal finding.

We run split-prefill on Qwen3.5-4B (32 layers, GDN) and
Falcon-H1-3B-Instruct (32 layers, Mamba) with four diagnostic tasks,
each isolating one capability:
\textbf{KV retrieve} (a small dictionary; $q$ asks for a value under a
key);
\textbf{list indexing} (a list; $q$ asks for the element at a position);
\textbf{language following} (an ``answer only in $L$'' instruction;
$q$ is in a different language; output language scored by a
language-ID classifier); and
\textbf{persona} ($c$ describes a persona; $q$ is neutral; style
match scored by an independent LLM judge \citep{zheng2023judging}).
We report means over three seeds with 95\% hierarchical bootstrap CIs.
Dataset sizes, prompts, full CIs across all four models, and the
layer-wise sweep modes are in
Appendices~\ref{app:datasets} and \ref{app:quant}.

The four tasks separate into two regimes whose ordering of
\textsc{kv-only} and \textsc{rec-only} is opposite
(Table~\ref{tab:main}, Fig.~\ref{fig:bars}).

\paragraph{Attention recalls items from prior context.} On KV
retrieve and list indexing, \textsc{kv-only} retains the bulk of the
full-model accuracy (64--98\% of baseline across both architectures)
while \textsc{rec-only} collapses to zero. The collapse persists in minimal versions of the task (short
dictionaries, short lists), ruling out a state-too-small explanation:
under split-prefill, recall of specific items from context
routes through the attention KV cache, while the recurrent channel
produces no retrieval.
\vspace{-0.6\baselineskip}
\paragraph{Recurrence controls the response mode.} On behavioural tasks
the ordering flips. \textsc{rec-only} retains 70--80\% of the
full-model language-following accuracy, while \textsc{kv-only} collapses
to $\sim$1\%, a 60--80$\times$ gap between the two channels. Persona
shows the same direction (\textsc{rec-only} 3--5$\times$ higher than
\textsc{kv-only}). The KV cache contains the language
instruction, yet the model does not act on it: the context establishes
a target \emph{mode of continuation} and only the recurrent state
controls it.

\begin{table}[t]
  \centering
  \small
  \setlength{\tabcolsep}{3pt}
  \begin{tabular}{l ccc ccc}
    \toprule
    & \multicolumn{3}{c}{\textbf{Qwen3.5-4B}}
    & \multicolumn{3}{c}{\textbf{Falcon-H1-3B}} \\
    \cmidrule(lr){2-4}\cmidrule(lr){5-7}
    \textbf{Task}
      & \textsc{full} & \textsc{rec} & \textsc{kv}
      & \textsc{full} & \textsc{rec} & \textsc{kv} \\
    \midrule
    KV retrieve    & 1.00 & 0.00 & \textbf{0.89}
                   & 1.00 & 0.00 & \textbf{0.64} \\
    List indexing  & 1.00 & 0.00 & \textbf{0.98}
                   & 0.96 & 0.00 & \textbf{0.83} \\
    Lang.\ follow  & 0.97 & \textbf{0.70} & 0.01
                   & 0.94 & \textbf{0.79} & 0.01 \\
    Persona        & 0.69 & \textbf{0.49} & 0.14
                   & 0.71 & \textbf{0.28} & 0.06 \\
    \bottomrule
  \end{tabular}
  \caption{Accuracy of the model under split-prefill on four diagnostic tasks. Top two rows: exact
  retrieval survives only the KV cache. Bottom two: language and
  persona transfer survive only the recurrent state.
  \textsc{rec}/\textsc{kv} = \textsc{rec-only}/\textsc{kv-only}.
  Details in Appendix~\ref{app:quant}.}
  \label{tab:main}
\end{table}

\section{Four Causal Case Studies}
\label{sec:demos}

We now visualise the dissociation through four causal case studies, ordered by the strength of their argument. The examples shown here are representative; full sample sizes, confidence intervals, and controls are in the appendix referenced at the end of each case study.

\subsection{Case Study \#1: Mixed Memories}
\label{sec:demo-swap}

We construct two contexts $c_A$ and $c_B$ that differ on two
independent properties: each carries its own small dictionary (so the
value stored under a given key is different in the two contexts) and
its own \emph{``answer only in $L$''} instruction (so the requested
language is also different). We prefill both contexts, then assemble one mixed
cache by taking the KV from $c_A$ together with the recurrent state
from $c_B$, and run a single shared query through this cache in one
forward pass (Figure~\ref{fig:stateswap}, Appendix~\ref{app:demo-swap}).

\begin{figure}[h!]
\begin{splitbox}
\plblA{ctx $c_A$} \hlA{dict $\{\textsc{moon}\!\to\!\texttt{A22},\dots\}$;\ \textit{answer in French}}\\
\plblB{ctx $c_B$} \hlB{dict $\{\textsc{moon}\!\to\!\texttt{B33},\dots\}$;\ \textit{answer in German}}\\
\plbl{cache} \textcolor{Acol}{$\mathcal{K}(c_A)$}\,$+$\,\textcolor{Bcol}{$\mathcal{R}(c_B)$}
\tcblower
\plbl{query} \textit{What is stored under the key \textsc{moon}?}\\[1pt]
\plbl{answers} \hlB{\emph{Der gespeicherte Wert ist}}\ \hlA{\texttt{A22}}.
\end{splitbox}
\vspace{-1em}
\caption{\textbf{Facts from KV, Style from Recurrence.}
Hybrid cache $\mathcal{K}(c_A) + \mathcal{R}(c_B)$ yields the value from $c_A$ and the language from $c_B$.}
\end{figure}

\noindent The model returns one fluent sentence in which the retrieved
value is the one stored in the KV-source context $c_A$, and the
language is the one requested by the recurrent-source context $c_B$.
The two properties are jointly correct on $\geq 99\%$ of items in
both models; same-source controls
($\mathcal{K}_A+\mathcal{R}_A$, $\mathcal{K}_B+\mathcal{R}_B$) match
\textsc{full}. Per-direction breakdown and CIs:
Appendix~\ref{app:demo-swap}.

\subsection{Case Study \#2: Recall, but Not Control}
\label{sec:demo-language}

\begin{figure}[h!]
\begin{splitbox}
\rowlbl{context}\emph{Answer only in French.}
\tcblower
\rowlbl{query}\emph{Hello, how are you?}\\[1pt]
\seclbl{answers}\\[1pt]
\hspace*{0.6em}\textsc{kv-only} $\to$ \emph{Hello!}\hfill
\textsc{rec-only} $\to$ \emph{Bonjour\,!}\\[1pt]
\noindent\textcolor{gray!55}{\rule{\linewidth}{0.3pt}}\\[1pt]
\rowlbl{query}\emph{What is your language for answer?}\\[1pt]
\seclbl{answers}\\[1pt]
\hspace*{0.6em}\textsc{kv-only} $\to$ \emph{French.}\hfill
\textsc{rec-only} $\to$ \emph{Français.}
\end{splitbox}
\vspace{-1em}
\caption{\textbf{Recall $\neq$ Control.}
KV-only names the requested language correctly but answers in English; rec-only reverses the pattern.}
\end{figure}

A starker test: ask the model under \textsc{kv-only} to \emph{name} the
requested language. The KV cache contains the relevant tokens, so the
model can retrieve them; but retrieving the name and acting on the
instruction are different things.

\noindent We pair every language setting with both a \emph{label}
query (what is the requested language?) and a \emph{behavioural} query
(a normal English question). Under \textsc{kv-only}, models name the
language correctly but answer in English; under \textsc{rec-only}, the
ordering inverts. The KV cache lets the model report what the
instruction was. To actually follow it, the recurrent state has to
be carried over from the context.

\subsection{Case Study \#3: Associative False Recall}
\label{sec:demo-drm}

If the recurrent state is not an addressable memory, what does it
preserve? We use a DRM-style \citep{roediger1995creating} list of words
strongly associated with an absent target.

\begin{figure}[h!]
\begin{splitbox}
\seclbl{context}\\[1pt]
\hspace*{0.6em}\emph{bed, rest, pillow, dream, blanket, tired, night, snore, nap}
\tcblower
\seclbl{query}\\[1pt]
\hspace*{0.6em}\emph{Was the exact word ``sleep'' in the list? Answer yes or no.} \hfill (gold: \emph{no})\\[1pt]
\seclbl{answers}\\[1pt]
\hspace*{0.6em}\textsc{kv-only} $\to$ \emph{No.}\hfill
\textsc{rec-only} $\to$ \emph{Yes.}
\end{splitbox}
\vspace{-1em}
\caption{\textbf{False Familiars.}
Rec-only falsely accepts an associated but absent target; KV-only correctly rejects.}
\end{figure}

\noindent Under \textsc{kv-only}, both models reject the trap; under
\textsc{rec-only}, false-yes rates rise sharply on the associated
list while a matched \emph{neutral-absent} control stays near floor.
The recurrent state preserves the \emph{semantic field} of seen words
rather than their identities: a target that was never tokenised in
context is nevertheless ``familiar'' enough to be reported as present.
Controls and per-seed rates: Appendix~\ref{app:demo-drm}.

\subsection{Case Study \#4: Conjunction Trap}
\label{sec:demo-conjunction}

A complementary failure mode is \emph{memory-conjunction errors}
\citep{reinitz1992memory}: features of two seen items recombine into an
unseen item that is mistaken for an original. If the recurrent state
stores feature-level rather than item-level evidence, it should fall
for this trap.

\begin{figure}[t!]
\begin{splitbox}
\seclbl{context}\\[1pt]
\hspace*{0.6em}\emph{sunflower, moonlight, blueberry, footprint, \dots}
\tcblower
\seclbl{query}\\[1pt]
\hspace*{0.6em}\emph{Was the exact word ``sunlight'' in the list?} \hfill (gold: \emph{no})\\[1pt]
\seclbl{answers}\\[1pt]
\hspace*{0.6em}\textsc{kv-only} $\to$ \emph{No.}\hfill
\textsc{rec-only} $\to$ \emph{Yes.}
\end{splitbox}
\vspace{-1em}
\caption{\textbf{Broken Binding.}
``sun-'' (from sunflower) and ``-light'' (from moonlight) both appear in the list; the compound ``sunlight'' does not.}
\end{figure}

\noindent Under \textsc{kv-only}, both models again reject the unseen
recombination. Under \textsc{rec-only}, false-yes rates rise sharply
on conjunction items but stay near floor on a \emph{single-part-absent}
control where only one of the two morphemes is present: the recurrence
signal is driven by the \emph{simultaneous} presence of both parts.
Component features of stored items survive in the recurrent state;
the binding that ties them together as a specific item is what gets
lost. Full controls: Appendix~\ref{app:demo-conjunction}.

\FloatBarrier
\section{Conclusion}
\label{sec:conclusion}

In a trained hybrid LM, the two components do not differ in how long
they retain past context. They differ in how the model accesses what
they hold. The KV cache works as an addressable store: the model can
pull specific items out of it. The recurrent state works as a
compressed prior that shapes language, persona, and semantic field of
generation. The claim is functional. It concerns what each channel
causally contributes to generation, not what each channel statically
encodes. State-swap makes the attribution causal at the level of a
single answer. The pattern parallels the explicit/implicit memory
distinction in cognitive psychology
\citep{tulving1985many,schacter1987implicit} and refines the
recall-deficiency story for state-space and linear-attention models
\citep{arora2024zoology,jelassi2024repeat,park2024can}: the recurrent
state should be read as a different kind of memory, and benchmarks
that score addressable lookup and behavioural conditioning together
cannot show this.

\section*{Limitations}

\paragraph{Functional, not static, content.} Our results characterise
what each channel \emph{causally drives} at inference, not what it
\emph{statically encodes}. We do not claim either channel is empty
of the other type of content. Demonstration~\ref{sec:demo-language}
is direct evidence to the contrary: the KV cache lets the model name
the requested language on demand, even when under \textsc{kv-only}
it cannot make the model answer in that language.

\paragraph{Probing left for future work.} Mapping the static
information content of attention and recurrent representations would
require trained probing classifiers on activations from each channel.
We leave this analysis to future work.

\paragraph{Nature of the intervention.} Split-prefill resets one
cache while preserving the other, which produces a partly
out-of-distribution combination: the two channels are normally
co-updated at every position, and resetting one breaks a coherence
the model relies on. Part of the rec-only / kv-only collapse could be
due to this incoherence, not to the channel itself being
uninformative. State-swap (Demo~\ref{sec:demo-swap}) addresses this
concern: both caches come from real prefills of well-formed contexts;
only the source contexts differ, so neither channel is in a
degenerate state. The same dissociation holds in this
closer-to-natural setting, which is harder to attribute to the
intervention itself.

\section*{Acknowledgements}
The study was implemented in the framework of the Basic Research Program at HSE University (HSE-BR-2025-025). We also acknowledge the computational resources of the HPC facilities at HSE University.


\appendix

\section{Implementation Details}
\label{app:impl}

\paragraph{Cache extraction.} We extract caches at the end of the context
prefill before any query token is processed. For Qwen3.5-4B we use the
HuggingFace \texttt{transformers} integration of Gated DeltaNet; for
Falcon-H1 we use the official HuggingFace release.

\section{Split-prefill: Detailed Procedure}
\label{app:split-prefill}

\paragraph{Notation.} A hybrid model has $L$ layers partitioned into
an attention set $\mathbb{A}$ and a recurrent set $\mathbb{R}$. Running
the model over a token sequence $c$ populates a memory state that
factorises across the two sets:
\[
  M(c) = \big(\underbrace{\{(K_\ell, V_\ell)\}_{\ell \in \mathbb{A}}}_{\mathcal{K}(c)},\
              \underbrace{\{(S_\ell, C_\ell)\}_{\ell \in \mathbb{R}}}_{\mathcal{R}(c)}\big),
\]
where $K_\ell, V_\ell \in \mathbb{R}^{h \times |c| \times d}$; $S_\ell$
is the layer's recurrent state (an SSM state for Mamba-style layers,
a matrix-valued state for delta-rule and gated-linear-attention
layers); and $C_\ell$ is the short causal convolution's state where
the architecture has one. $C_\ell$ is treated as part of the recurrent
channel throughout: carried with $S_\ell$, dropped with it. The KV
channel grows with $|c|$ and is content-addressable; the recurrent
channel is a fixed-size summary. This asymmetry is what split-prefill
probes.

\paragraph{The three conditions.} Every example is a pair $(c, q)$: a
context carrying the information under test and a query identical
across conditions. We run the model over $c$ alone, retain $M(c)$,
construct a fresh memory state, populate one or both channels, and
then run over $q$:
\begin{itemize}\setlength\itemsep{1pt}
  \item \textsc{full}: $M \leftarrow (\mathcal{K}(c),\ \mathcal{R}(c))$; no intervention.
  \item \textsc{kv-only}: $M \leftarrow (\mathcal{K}(c),\ \emptyset)$; the recurrent state is zeroed.
  \item \textsc{rec-only}: $M \leftarrow (\emptyset,\ \mathcal{R}(c))$; the KV cache is dropped.
\end{itemize}
Three properties matter. First, the context is prefilled once; the
ablated conditions reuse the state from the same forward pass over $c$
and do not re-run the model on a truncated context. Second, the query
is byte-identical across conditions and tokenised once. Third,
positions follow the retained KV: under \textsc{full} and
\textsc{kv-only}, $q$ begins at position $|c|$; under \textsc{rec-only}
the KV is empty so $q$ begins at $0$. This is deliberate. A dropped
cache has no positions to occupy, and padding with zero keys to
equalise positions would introduce a third, uncontrolled condition,
since zero keys are still attended to.

\paragraph{Constructing $c$ and $q$.} The split must land inside a
single user turn, so that no template markup falls between $c$ and $q$
and the boundary is not a role change. We render the whole conversation
once, $T = \mathrm{template}(\{\text{user}: u_1\Vert u_2\},\ \text{add\_generation\_prompt})$,
locate the first occurrence of $u_1$ in $T$, and cut immediately
after it. Cutting a rendered string rather than concatenating two
separately rendered fragments guarantees that $c\Vert q$ is exactly
what the model would see without intervention, so the \textsc{full}
condition coincides with ordinary inference. For reasoning models we
append an empty \texttt{<think></think>} block to the assistant opener;
without it every condition answers in English reasoning prose, which
destroys the language metric.

\paragraph{The recurrent-state carry hazard.} Most public
implementations of hybrid layers only consume a restored recurrent
state on the single-token decode path. A common gate reads
\begingroup\small
\begin{verbatim}
use_precomputed_states = cache is not None
    and cache.has_previous_state(l)
    and seq_len == 1
\end{verbatim}
\endgroup
\normalsize
\noindent so when $q$ is fed as one multi-token chunk the restored
$S_\ell$ is silently ignored and the scan restarts from zero. No error
is raised. \textsc{rec-only} then measures a model with no context at
all, and its collapse to chance is easily misread as evidence that the
recurrent channel carries nothing. Two equivalent remedies, chosen per
model: patch the mixer to pass the restored state as an
\texttt{initial\_state} argument to the chunked kernel, or feed $q$
token-by-token so the stock gate admits it.

\paragraph{Verification protocol (mandatory per model).} Because the
failure above is silent, we verify before recording any number that
the restored state actually influences the output. Holding everything
else fixed:
\[
  \Delta \;=\; \big\|\mathrm{logits}(q \mid \mathcal{R}(c))
                    \;-\; \mathrm{logits}(q \mid \mathbf{0})\big\|_\infty.
\]
$\Delta \approx 0$ proves the state never reached the model and
invalidates every \textsc{rec-only} and crossed-cache number for that
configuration; the run is discarded rather than reported. We observed
exactly this on one model under chunked prefill ($\Delta = 0.0000$),
fixed by the token-wise remedy ($\Delta = 12.1$). Two cautions. First,
healthy controls do not substitute for this check: in the crossed-cache
experiment the same-source controls can look perfectly healthy while
the state is dropped, because with both channels from a single context
the KV alone explains the outputs. Second, verification must run
inside the exact code path used for the reported numbers, including
the specific model class and library version.

\paragraph{Crossed cache (state-swap).} The causal variant pairs
contexts $c_A, c_B$ differing in both retrieved content and instructed
response mode, assembles $M \leftarrow (\mathcal{K}(c_A),\ \mathcal{R}(c_B))$,
and runs the shared query. Same-source assemblies serve as controls and
bound achievable accuracy. We assert at run time that $q$ renders
identically for both. For implementations taking an explicit starting
position, the position passed with $q$ is the length of the KV-source
context, keeping positions consistent with the restored keys.

\paragraph{Decoding and scoring.} Greedy (argmax), no sampling, stopping
at EOS or after 8 new tokens for retrieval (single-word answers) and
48 for free-text. Retrieval is scored by exact match of the first
alphabetic word. Language is scored with an off-the-shelf identifier
restricted to the target set, so a confident prediction outside it
counts as failure rather than being silently mapped. Persona is scored
by an instruction-tuned LLM judge emitting a binary verdict, held
fixed across conditions and models. Since all conditions share query,
prompts, and decoder, differences cannot arise from decoding.

\paragraph{Per-model notes.} A per-model summary records where the
state is stored, which carry remedy was used, and the verified
$\Delta$. Two practical hazards. Several checkpoints ship custom
modelling code pinned to an older library version, and running them
under a newer one can fail silently, producing fluent-looking but
degenerate output in every condition rather than an exception. Each
such model is therefore run under the library version its code
targets, with raw generations inspected before scoring. A renamed
dtype or position argument across versions fails silently as well.
Both hazards are caught by the verification above combined with a
visual check of the control generations.

\section{Datasets}
\label{app:datasets}

All datasets are programmatically generated.

\paragraph{KV retrieve.} A dictionary with 15--30 entries; keys and
values are short ASCII tokens unlikely to overlap with each other. The
query asks for the value under a single key. Gold: the exact value as a
string. The lower bound on entry count ensures retrieval is non-trivial;
the upper bound keeps total length manageable.

\paragraph{List indexing.} A Python-style list of 4--10 short tokens.
The query asks for the element at a numeric position. Gold: the exact
list element.

\paragraph{Language following.} 18 target languages
(\textsc{en, fr, de, es, it, pt, nl, ru, pl, tr, uk, sv, da, ja, zh, ko,
hi, id}). The context contains the instruction ``Answer only in $L$'';
the query is a meaning-preserving English question. We score with
\texttt{fastText} \texttt{lid.176} \citep{joulin2017bag}, taking
the highest-probability language label of the generation.

\paragraph{Persona.} 15 persona descriptions
(pirate, cosmonaut, wizard, samurai, knight, vampire, alien, surfer,
caveman, sage, robot, detective, ninja, shaman, cowboy) paired with 15
neutral questions. An independent judge LM (Qwen2.5-72B-Instruct,
specifically chosen to be in a different family than the tested
models) returns a $[0,1]$ persona-match score.

\paragraph{Association trap.} 100 DRM-style associated lists of 8--10
words each, drawn from the Roediger--McDermott norms; plus 100 matched
neutral controls (unrelated words) and 100 positive-present controls
(target word actually placed in the list).

\paragraph{Conjunction trap.} 200 conjunction items
(e.g.\ \emph{sunflower} + \emph{moonlight} $\to$ probe
\emph{sunlight}). Three matched controls: \emph{neutral-absent}
(target unrelated to any list item), \emph{single-part-absent} (only
one component morpheme present in the list), and \emph{positive-present}
(the conjunction item itself in the list).

\paragraph{Crossed dictionary + language (Demo 1).} 100 crossed examples
per direction ($\mathcal{K}_A+\mathcal{R}_B$ and the mirror) plus 200
matched same-source controls ($\mathcal{K}_A+\mathcal{R}_A$ and
$\mathcal{K}_B+\mathcal{R}_B$). $L_A, L_B$ drawn from
\{fr, de, es, it, ja, ko\}.

\section{Quantitative Dissociation: Full Results}
\label{app:quant}

Table~\ref{tab:app-main-all} reproduces Table~\ref{tab:main} across
all four evaluated models with 95\% hierarchical bootstrap CIs
(3 seeds, 10{,}000 replicates, resampling seeds then examples within
seed). The two recurrent-only retrieval cells are 0/600 on every
seed for every model. The two recurrent-only behavioural cells
substantially exceed kv-only on every model with the single
exception of Falcon-H1-7B persona (kv-only $0.29$, rec-only $0.25$;
paired bootstrap $-3.8$~p.p.\ $[-6.9, -0.6]$, marked $^\dagger$).
Across both families, kv-only retrieval accuracy improves monotonically
with model size while the recurrent-only collapse persists, preserving
the dissociation.

Table~\ref{tab:app-layer} reports Qwen-specific layer-wise sweep
modes: \textsc{lower-kv+full-rec} keeps the attention KV cache only
for the bottom half of attention layers and the recurrent state in
full; \textsc{full-kv+lower-rec} keeps the KV cache in full and the
recurrent state only for the lower half of GDN layers. Retrieval is
preserved only when KV is kept in full; behavioural transfer is
preserved whenever the recurrent state is kept in full. These modes
were not implemented for Falcon-H1.

\paragraph{Persona judge.} Qwen2.5-72B-Instruct with three samples
per response. We use no model from the tested families (Qwen3.5 or
Falcon-H1) to avoid self-judging.

\begin{table*}[ht]
  \centering
  \small
  \setlength{\tabcolsep}{6pt}
  \begin{tabular}{l l ccc}
    \toprule
    \textbf{Model} & \textbf{Task} & \textsc{full} & \textsc{rec-only} & \textsc{kv-only} \\
    \midrule
    Qwen3.5-4B   & KV retrieve  & 1.00 [0.99, 1.00]  & 0.00 [0.00, 0.00] & 0.89 [0.86, 0.92] \\
                 & List index   & 1.00 [0.99, 1.00]  & 0.00 [0.00, 0.00] & 0.98 [0.96, 0.99] \\
                 & Lang.\ foll. & 0.97 [0.96, 0.98]    & 0.70 [0.67, 0.72] & 0.01 [0.01, 0.02] \\
                 & Persona      & 0.69 [0.66, 0.72]    & 0.49 [0.46, 0.53] & 0.14 [0.12, 0.16] \\
    \midrule
    Qwen3.5-9B   & KV retrieve  & 1.00 [0.99, 1.00]  & 0.00 [0.00, 0.00] & 0.94 [0.92, 0.96] \\
                 & List index   & 1.00 [0.99, 1.00]  & 0.00 [0.00, 0.00] & 0.99 [0.98, 1.00] \\
                 & Lang.\ foll. & 0.96 [0.94, 0.97]    & 0.62 [0.60, 0.65] & 0.01 [0.00, 0.01] \\
                 & Persona      & 0.74 [0.72, 0.77]    & 0.63 [0.60, 0.66] & 0.08 [0.07, 0.10] \\
    \midrule
    Falcon-H1-3B & KV retrieve  & 1.00 [0.99, 1.00]  & 0.00 [0.00, 0.00] & 0.64 [0.60, 0.68] \\
                 & List index   & 0.96 [0.94, 0.97]    & 0.00 [0.00, 0.00] & 0.83 [0.79, 0.86] \\
                 & Lang.\ foll. & 0.94 [0.93, 0.95]    & 0.79 [0.77, 0.82] & 0.01 [0.00, 0.01] \\
                 & Persona      & 0.71 [0.68, 0.73]    & 0.28 [0.25, 0.31] & 0.06 [0.05, 0.08] \\
    \midrule
    Falcon-H1-7B & KV retrieve  & 1.00 [1.00, 1.00] & 0.00 [0.00, 0.00] & 0.99 [0.98, 1.00] \\
                 & List index   & 1.00 [1.00, 1.00] & 0.00 [0.00, 0.00] & 0.98 [0.96, 0.99] \\
                 & Lang.\ foll. & 0.92 [0.91, 0.93]    & 0.61 [0.59, 0.64] & 0.05 [0.04, 0.06] \\
                 & Persona      & 0.75 [0.72, 0.78]    & 0.25 [0.22, 0.28] & 0.29 [0.26, 0.32]$^\dagger$ \\
    \midrule
    Qwen3.5-27B$^\ddagger$ & KV retrieve  & 1.00 [1.00, 1.00] & 0.00 [0.00, 0.00] & 0.99 [0.98, 1.00] \\
                 & List index   & 1.00 [1.00, 1.00] & 0.00 [0.00, 0.00] & 1.00 [1.00, 1.00] \\
                 & Lang.\ foll. & 0.99 [0.98, 1.00]   & 0.83 [0.79, 0.86] & 0.08 [0.06, 0.11] \\
                 & Persona      & 0.74 [0.69, 0.78]    & 0.66 [0.61, 0.71] & 0.31 [0.26, 0.36] \\
    \midrule
    Falcon-H1-34B$^\ddagger$ & KV retrieve  & 1.00 [1.00, 1.00] & 0.00 [0.00, 0.00] & 0.85 [0.80, 0.90] \\
                 & List index   & 1.00 [1.00, 1.00] & 0.02 [0.00, 0.04] & 0.96 [0.93, 0.99] \\
                 & Lang.\ foll. & 0.96 [0.94, 0.98]    & 0.41 [0.36, 0.46] & 0.25 [0.21, 0.29] \\
                 & Persona      & 0.80 [0.75, 0.84]    & 0.18 [0.14, 0.23] & 0.33 [0.28, 0.39]$^\dagger$ \\
    \midrule
    Kimi-Linear-48B$^\ast$  & KV retrieve  & 0.99 & 0.00 & 0.94 \\
                 & List index   & 0.95 & 0.00 & 0.91 \\
                 & Lang.\ foll. & 0.94 & 0.77 & 0.00 \\
                 & Persona      & 0.99 & 0.80 & 0.14 \\
    \midrule
    Olmo-Hybrid-7B$^\ast$   & KV retrieve  & 0.99 & 0.00 & 0.99 \\
                 & List index   & 0.98 & 0.00 & 0.91 \\
                 & Lang.\ foll. & 0.97 & 0.72 & 0.01 \\
                 & Persona      & 0.99 & 0.94 & 0.08 \\
    \midrule
    Jamba2-3B$^\ast$        & KV retrieve  & 0.59 & 0.00 & 0.77 \\
                 & List index   & 0.79 & 0.00 & 0.40 \\
                 & Lang.\ foll. & 0.63 & 0.55 & 0.07 \\
                 & Persona      & 0.98 & 0.75 & 0.36 \\
    \bottomrule
  \end{tabular}
  \caption{Main split-prefill results across all evaluated models.
  Rows without markers: means over 3 seeds with 95\% hierarchical
  bootstrap CIs in brackets. $n{=}200$ per seed for retrieval, 450
  for language, 225 for persona. $^\dagger$Cells where \textsc{kv-only}
  exceeds \textsc{rec-only} on a behavioural task; see text.
  $^\ddagger$Single-seed extension (seed 42 only); intervals are
  within-seed bootstraps and do not quantify across-seed variability.
  $^\ast$Additional architectures; point estimates only (bootstrap
  CIs not computed for these runs).}
  \label{tab:app-main-all}
\end{table*}

\begin{table}[ht]
  \centering
  \footnotesize
  \setlength{\tabcolsep}{3pt}
  \begin{tabular}{l cc}
    \toprule
    \textbf{Task} & \begin{tabular}{@{}c@{}}lower KV\\+ full rec\end{tabular}
                  & \begin{tabular}{@{}c@{}}full KV\\+ lower rec\end{tabular} \\
    \midrule
    KV retrieve   & 0.02 [0.01, 0.03] & 1.00 [0.99, 1.00] \\
    List indexing & 0.01 [0.00, 0.02] & 1.00 [0.99, 1.00] \\
    Lang.\ follow & 0.95 [0.94, 0.96] & 0.71 [0.68, 0.73] \\
    Persona       & 0.68 [0.65, 0.71] & 0.65 [0.62, 0.68] \\
    \bottomrule
  \end{tabular}
  \caption{Layer-wise sweep on Qwen3.5-4B. The two modes dissociate
  layer-wise as well as channel-wise: retrieval requires KV in full,
  behavioural transfer requires recurrent state in full.}
  \label{tab:app-layer}
\end{table}

\section{Demonstration 1 --- State-swap}
\label{app:demo-swap}

\begin{figure*}[ht]
  \centering
  \includegraphics[width=\linewidth]{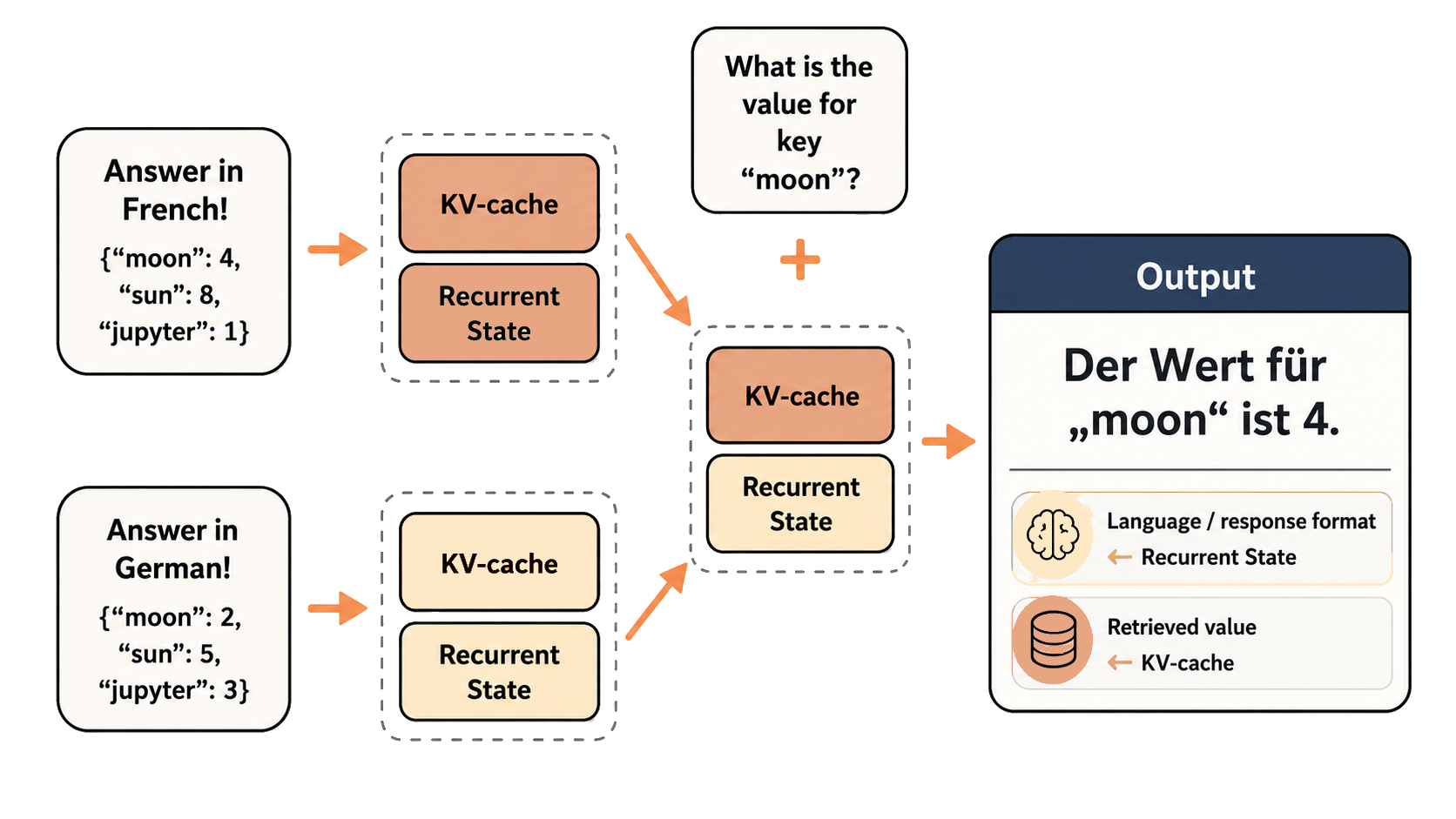}
  \caption{State-swap construction. Two contexts $c_A$ and $c_B$ each
  pair a dictionary with a language instruction; prefilling each yields
  its own KV cache and recurrent state. We assemble a hybrid cache (KV
  from $c_A$ + recurrent state from $c_B$), feed a shared query, and
  read off which channel each property of the answer follows. In the
  illustrated cross, the answer language (German) follows the recurrent
  source $c_B$ while the dictionary value is retrieved from the KV
  source $c_A$.}
  \label{fig:stateswap}
\end{figure*}

\paragraph{Construction.} 100 crossed examples per direction
($\mathcal{K}_A+\mathcal{R}_B$, $\mathcal{K}_B+\mathcal{R}_A$;
$n{=}200$ total crossed per seed) plus 200 matched same-source
controls ($\mathcal{K}_A+\mathcal{R}_A$, $\mathcal{K}_B+\mathcal{R}_B$).
Match criterion for \emph{value$\to$KV}: case-insensitive substring
match of the KV-source dictionary value in the output. Match criterion
for \emph{lang$\to$rec}: \texttt{fastText} \texttt{lid.176} identifies
the recurrent-source language as the top-probability label.

\paragraph{Results.} Table~\ref{tab:app-demo-swap} reports both
crossing directions for each model. On Qwen-4B and Qwen-9B every
crossed example is jointly correct. Falcon-H1-3B is near-perfect
($\geq0.99$ both-correct in either direction). Falcon-H1-7B retains
perfect value retrieval but loses some language behaviour
($\approx0.87$--$0.89$ both-correct), consistent with its weaker
recurrent-only language-following score in
Table~\ref{tab:app-main-all}. Same-source controls
($\mathcal{K}_A+\mathcal{R}_A$ and $\mathcal{K}_B+\mathcal{R}_B$)
reach $1.00$ both-correct for Qwen-4B, Qwen-9B and Falcon-H1-7B;
Falcon-H1-3B controls are also near-perfect ($\sim0.99$).

\begin{table*}[ht]
  \centering
  \small
  \setlength{\tabcolsep}{6pt}
  \begin{tabular}{l l ccc}
    \toprule
    \textbf{Model} & \textbf{Crossed state} &
      value$\to$KV & lang$\to$rec & both \\
    \midrule
    Qwen3.5-4B   & $\mathcal{K}_A+\mathcal{R}_B$ &
      1.00 [1.00, 1.00] & 1.00 [1.00, 1.00] & 1.00 [1.00, 1.00] \\
                 & $\mathcal{K}_B+\mathcal{R}_A$ &
      1.00 [1.00, 1.00] & 1.00 [1.00, 1.00] & 1.00 [1.00, 1.00] \\
    \midrule
    Qwen3.5-9B   & $\mathcal{K}_A+\mathcal{R}_B$ &
      1.00 [1.00, 1.00] & 1.00 [1.00, 1.00] & 1.00 [1.00, 1.00] \\
                 & $\mathcal{K}_B+\mathcal{R}_A$ &
      1.00 [1.00, 1.00] & 1.00 [1.00, 1.00] & 1.00 [1.00, 1.00] \\
    \midrule
    Falcon-H1-3B & $\mathcal{K}_A+\mathcal{R}_B$ &
      1.00 [0.99, 1.00]  & 1.00 [0.99, 1.00]  & 0.99 [0.98, 1.00] \\
                 & $\mathcal{K}_B+\mathcal{R}_A$ &
      1.00 [0.99, 1.00]  & 0.99 [0.98, 1.00]   & 0.99 [0.98, 1.00] \\
    \midrule
    Falcon-H1-7B & $\mathcal{K}_A+\mathcal{R}_B$ &
      1.00 [1.00, 1.00] & 0.89 [0.85, 0.93]    & 0.89 [0.85, 0.93] \\
                 & $\mathcal{K}_B+\mathcal{R}_A$ &
      1.00 [1.00, 1.00] & 0.87 [0.83, 0.91]    & 0.87 [0.83, 0.91] \\
    \midrule
    Qwen3.5-27B$^\ddagger$ & $\mathcal{K}_A+\mathcal{R}_B$ &
      1.00 [1.00, 1.00] & 0.96 [0.92, 0.99]    & 0.96 [0.92, 0.99] \\
                 & $\mathcal{K}_B+\mathcal{R}_A$ &
      1.00 [1.00, 1.00] & 0.95 [0.90, 0.99]    & 0.95 [0.90, 0.99] \\
    \midrule
    Falcon-H1-34B$^\ddagger$ & $\mathcal{K}_A+\mathcal{R}_B$ &
      1.00 [1.00, 1.00] & 0.72 [0.63, 0.81]    & 0.72 [0.63, 0.81] \\
                 & $\mathcal{K}_B+\mathcal{R}_A$ &
      1.00 [1.00, 1.00] & 0.71 [0.62, 0.80]    & 0.71 [0.62, 0.80] \\
    \midrule
    Kimi-Linear-48B$^\ast$   & $\mathcal{K}_A+\mathcal{R}_B$ &
      1.00 & 1.00 & 1.00 \\
                 & $\mathcal{K}_B+\mathcal{R}_A$ &
      1.00 & 0.98  & 0.98 \\
    \midrule
    Olmo-Hybrid-7B$^\ast$    & $\mathcal{K}_A+\mathcal{R}_B$ &
      1.00 & 0.96 & 0.96 \\
                 & $\mathcal{K}_B+\mathcal{R}_A$ &
      1.00 & 0.94 & 0.94 \\
    \midrule
    Jamba2-3B$^\ast$         & $\mathcal{K}_A+\mathcal{R}_B$ &
      1.00 & 0.92 & 0.92 \\
                 & $\mathcal{K}_B+\mathcal{R}_A$ &
      1.00 & 0.90 & 0.90 \\
    \bottomrule
  \end{tabular}
  \caption{State-swap (Demo 1). Each row is one crossing direction.
  Unmarked rows: pooled across 3 seeds, $n{=}100$ examples per seed
  per direction ($n{=}300$ total per row).
  $^\ddagger$Single-seed extension (seed 42 only); intervals are
  within-seed bootstraps.
  $^\ast$Additional architectures; point estimates only. For Jamba2-3B
  the same-source controls reach only $0.86$--$0.89$ both-correct.}
  \label{tab:app-demo-swap}
\end{table*}

\section{Demonstration 2 --- Label vs.\ behavioural language}
\label{app:demo-language}

\paragraph{Construction.} A 20-language self-report probe of the form
\emph{``Answer only in $L$. What is your language for answer? Answer
in one word''} ($n{=}20$ per seed, 3 seeds). \emph{English label}:
the output is the English name of $L$ (e.g.\ \emph{Russian}); this
exposes literal label retrieval from the KV cache. \emph{Target-language
label}: the output is the language name in $L$ itself (e.g.\
\emph{Russkij} for Russian or \emph{Deutsch} for German); this requires
the model to also enact the language behaviour.

\paragraph{Results.} Table~\ref{tab:app-demo-language} shows both
scores together. Qwen3.5-9B is the cleanest case:
\textsc{kv-only} names the language in English $90\%$ of the time
while almost never speaking it ($0.05$ target-label), and
\textsc{rec-only} reverses the pattern ($0.10$ English / $0.95$ target).
Qwen3.5-4B and Falcon-H1-7B show the same direction. Falcon-H1-3B is
weaker on this specific self-report prompt despite strong ordinary
language-following accuracy ($0.79$ \textsc{rec-only} vs.\ $0.01$
\textsc{kv-only} on the $450$-item dataset, Table~\ref{tab:main});
the broader argument rests on the larger language-following dataset
and on the three other models here.

\begin{table*}[ht]
  \centering
  \small
  \setlength{\tabcolsep}{4pt}
  \begin{tabular}{l l cc}
    \toprule
    \textbf{Model} & \textbf{Mode} & Eng.\ label & Target-lang.\ label \\
    \midrule
    Qwen3.5-4B   & \textsc{full}     & 0.15 [0.07, 0.25]    & 0.90 [0.82, 0.97] \\
                 & \textsc{rec-only} & 0.05 [0.00, 0.12]    & 0.95 [0.88, 1.00] \\
                 & \textsc{kv-only}  & 0.50 [0.38, 0.63]    & 0.05 [0.00, 0.12] \\
    \midrule
    Qwen3.5-9B   & \textsc{full}     & 0.05 [0.00, 0.12]    & 1.00 [1.00, 1.00] \\
                 & \textsc{rec-only} & 0.10 [0.03, 0.18]    & 0.95 [0.88, 1.00] \\
                 & \textsc{kv-only}  & 0.90 [0.82, 0.97]    & 0.05 [0.00, 0.12] \\
    \midrule
    Falcon-H1-3B & \textsc{full}     & 0.05 [0.00, 0.12]    & 0.85 [0.75, 0.93] \\
                 & \textsc{rec-only} & 0.00 [0.00, 0.00]    & 0.10 [0.03, 0.18] \\
                 & \textsc{kv-only}  & 0.55 [0.43, 0.67]    & 0.40 [0.28, 0.52] \\
    \midrule
    Falcon-H1-7B & \textsc{full}     & 0.10 [0.03, 0.18]    & 0.95 [0.88, 1.00] \\
                 & \textsc{rec-only} & 0.05 [0.00, 0.12]    & 0.50 [0.37, 0.62] \\
                 & \textsc{kv-only}  & 1.00 [1.00, 1.00] & 0.05 [0.00, 0.12] \\
    \midrule
    Qwen3.5-27B$^\ddagger$ & \textsc{full}     & 0.05 & 1.00 \\
                 & \textsc{rec-only} & 0.05 & 0.95 \\
                 & \textsc{kv-only}  & 0.85 & 0.20 \\
    \midrule
    Falcon-H1-34B$^\ddagger$ & \textsc{full}     & 0.15 & 0.85 \\
                 & \textsc{rec-only} & 0.65 & 0.10 \\
                 & \textsc{kv-only}  & 0.35 & 0.05 \\
    \bottomrule
  \end{tabular}
  \caption{Self-report probe (Demo 2). Fraction of responses that
  contain the English language name (col.\ 3) or the language name
  in the target language (col.\ 4). Unmarked rows: $n{=}20$ per seed,
  3 seeds, 95\% bootstrap CIs. $^\ddagger$Single-seed extension
  (seed 42 only, one example per target language); point estimates
  shown without intervals.}
  \label{tab:app-demo-language}
\end{table*}

\section{Demonstration 3 --- Associative (DRM) trap}
\label{app:demo-drm}

\paragraph{Construction.} Per seed: 100 associated lists
(Roediger--McDermott norms; 8--10 words each, target word absent),
100 matched neutral lists (target unrelated to list contents,
absent), 100 positive-present lists (target word actually placed at
a random position). The query in every condition is \emph{``Was the
exact word `$w$' in the list?''}. False-positive rate (FPR) is the
fraction of \emph{yes} responses on absent-target conditions; true
positive rate (TPR) is the fraction of \emph{yes} on the present
condition.

\paragraph{Results.} Table~\ref{tab:app-demo-drm} shows the three
rates per model and mode. \textsc{rec-only} drives a large
associated-absent FPR on every model (Qwen-4B $0.87$, Qwen-9B $0.89$,
Falcon-3B $0.45$, Falcon-7B $0.99$). For three of four models the
neutral control stays at or below $0.10$, so the false yes is
specifically driven by the associative cue rather than a blanket
yes-bias. Falcon-H1-7B is the exception: its \textsc{rec-only}
neutral FPR is $0.18$ and \textsc{kv-only} associated/neutral are
both above $0.55$, indicating broad false-membership susceptibility
rather than a clean associative effect; the rec-only associated
rate still exceeds the rec-only neutral rate by $0.81$~p.p., but the
interpretation should be more guarded for this model.

\begin{table*}[ht]
  \centering
  \small
  \setlength{\tabcolsep}{6pt}
  \begin{tabular}{l l ccc}
    \toprule
    \textbf{Model} & \textbf{Mode} &
      assoc-absent FPR & neutral-absent FPR & present TPR \\
    \midrule
    Qwen3.5-4B   & \textsc{full}     & 0.00 [0.00, 0.00]  & 0.00 [0.00, 0.00] & 0.97 [0.94, 0.99] \\
                 & \textsc{rec-only} & 0.87 [0.82, 0.91]  & 0.10 [0.07, 0.14] & 0.98 [0.96, 1.00] \\
                 & \textsc{kv-only}  & 0.00 [0.00, 0.00]  & 0.00 [0.00, 0.00] & 0.71 [0.65, 0.77] \\
    \midrule
    Qwen3.5-9B   & \textsc{full}     & 0.00 [0.00, 0.00]  & 0.00 [0.00, 0.00] & 1.00 [1.00, 1.00] \\
                 & \textsc{rec-only} & 0.89 [0.84, 0.93]  & 0.01 [0.00, 0.02] & 0.93 [0.90, 0.96] \\
                 & \textsc{kv-only}  & 0.01 [0.00, 0.03]  & 0.00 [0.00, 0.00] & 0.86 [0.81, 0.92] \\
    \midrule
    Falcon-H1-3B & \textsc{full}     & 0.00 [0.00, 0.00]  & 0.00 [0.00, 0.00] & 1.00 [0.99, 1.00] \\
                 & \textsc{rec-only} & 0.45 [0.38, 0.52]  & 0.00 [0.00, 0.01] & 0.72 [0.66, 0.77] \\
                 & \textsc{kv-only}  & 0.05 [0.02, 0.09]  & 0.01 [0.00, 0.02] & 0.71 [0.64, 0.77] \\
    \midrule
    Falcon-H1-7B & \textsc{full}     & 0.00 [0.00, 0.00]  & 0.00 [0.00, 0.00] & 1.00 [1.00, 1.00] \\
                 & \textsc{rec-only} & 0.99 [0.97, 1.00] & 0.18 [0.14, 0.23] & 0.99 [0.98, 1.00] \\
                 & \textsc{kv-only}  & 0.60 [0.55, 0.66]  & 0.56 [0.49, 0.62] & 1.00 [0.99, 1.00] \\
    \midrule
    Qwen3.5-27B$^\ddagger$ & \textsc{full}     & 0.00 [0.00, 0.00]  & 0.00 [0.00, 0.00] & 1.00 [1.00, 1.00] \\
                 & \textsc{rec-only} & 0.29 [0.20, 0.38]  & 0.01 [0.00, 0.03] & 0.48 [0.38, 0.58] \\
                 & \textsc{kv-only}  & 0.29 [0.20, 0.38]  & 0.20 [0.13, 0.28] & 1.00 [1.00, 1.00] \\
    \midrule
    Falcon-H1-34B$^\ddagger$ & \textsc{full}     & 0.00 [0.00, 0.00]  & 0.00 [0.00, 0.00] & 1.00 [1.00, 1.00] \\
                 & \textsc{rec-only} & 0.72 [0.63, 0.81]  & 0.00 [0.00, 0.00] & 1.00 [1.00, 1.00] \\
                 & \textsc{kv-only}  & 0.32 [0.23, 0.41]  & 0.11 [0.05, 0.18] & 0.86 [0.79, 0.92] \\
    \bottomrule
  \end{tabular}
  \caption{DRM trap (Demo 3). Three rates per (model, mode):
  associated-absent false-positive (the trap), neutral-absent
  false-positive (yes-bias control), and present true-positive
  (verifies the model can affirm true membership). Unmarked rows:
  $n{=}100$ per condition per seed, 3 seeds, 95\% bootstrap CIs.
  $^\ddagger$Single-seed extension (seed 42 only); intervals are
  within-seed bootstraps.}
  \label{tab:app-demo-drm}
\end{table*}

\section{Demonstration 4 --- Memory-conjunction trap}
\label{app:demo-conjunction}

\paragraph{Construction.} Per seed: 200 conjunction items, each a
list of 8--12 single-token words whose subset includes both morphemes
of a target compound (e.g.\ \emph{sunflower} and \emph{moonlight} for
the probe \emph{sunlight}). Four matched controls: \emph{neutral-absent}
(target unrelated to any list item), \emph{left-only} and
\emph{right-only} (only one of the two morphemes present in the list),
and \emph{positive-present} (the conjunction itself in the list).

\paragraph{Results.} Table~\ref{tab:app-demo-conj} reports all five
rates per (model, mode). On all four models the \textsc{rec-only}
conjunction-absent rate exceeds both single-part-absent rates:
Qwen-4B $0.32$ vs.\ $0.18/0.05$; Qwen-9B $0.17$ vs.\ $0.08/0.03$;
Falcon-3B $0.15$ vs.\ $0.06/0.04$; Falcon-7B $0.75$ vs.\ $0.61/0.48$.
The signal therefore amplifies when both component morphemes are
simultaneously present, beyond what either part alone produces.
Falcon-H1-7B again shows broad recurrent-only false-positive
susceptibility (neutral $0.68$, both single-parts above $0.48$);
the conjunction effect persists relatively, but absolute rates are
inflated.

\begin{table*}[ht]
  \centering
  \footnotesize
  \setlength{\tabcolsep}{4pt}
  \begin{tabular}{l l ccccc}
    \toprule
    \textbf{Model} & \textbf{Mode} &
      conj.\ FPR & neutral FPR & left-only FPR & right-only FPR & present TPR \\
    \midrule
    Qwen3.5-4B   & \textsc{full}     & 0.00 [0.00, 0.00] & 0.00 [0.00, 0.00] & 0.00 [0.00, 0.00] & 0.00 [0.00, 0.00] & 1.00 [1.00, 1.00] \\
                 & \textsc{rec-only} & 0.32 [0.28, 0.36] & 0.10 [0.07, 0.12] & 0.18 [0.15, 0.22] & 0.05 [0.04, 0.07] & 0.53 [0.49, 0.57] \\
                 & \textsc{kv-only}  & 0.00 [0.00, 0.00] & 0.00 [0.00, 0.00] & 0.00 [0.00, 0.00] & 0.00 [0.00, 0.00] & 0.79 [0.75, 0.82] \\
    \midrule
    Qwen3.5-9B   & \textsc{full}     & 0.00 [0.00, 0.00] & 0.00 [0.00, 0.00] & 0.00 [0.00, 0.00] & 0.00 [0.00, 0.00] & 1.00 [1.00, 1.00] \\
                 & \textsc{rec-only} & 0.17 [0.13, 0.20] & 0.03 [0.02, 0.05] & 0.08 [0.05, 0.10] & 0.03 [0.01, 0.04] & 0.27 [0.23, 0.30] \\
                 & \textsc{kv-only}  & 0.06 [0.04, 0.08] & 0.00 [0.00, 0.01] & 0.04 [0.03, 0.06] & 0.02 [0.01, 0.03] & 0.77 [0.73, 0.81] \\
    \midrule
    Falcon-H1-3B & \textsc{full}     & 0.02 [0.00, 0.03] & 0.00 [0.00, 0.01] & 0.00 [0.00, 0.00] & 0.00 [0.00, 0.00] & 1.00 [0.99, 1.00] \\
                 & \textsc{rec-only} & 0.15 [0.11, 0.20] & 0.01 [0.00, 0.02] & 0.06 [0.04, 0.09] & 0.04 [0.02, 0.06] & 0.30 [0.24, 0.35] \\
                 & \textsc{kv-only}  & 0.01 [0.00, 0.02] & 0.04 [0.02, 0.06] & 0.00 [0.00, 0.00] & 0.00 [0.00, 0.01] & 0.65 [0.61, 0.69] \\
    \midrule
    Falcon-H1-7B & \textsc{full}     & 0.01 [0.00, 0.02] & 0.00 [0.00, 0.00] & 0.00 [0.00, 0.01] & 0.00 [0.00, 0.00] & 1.00 [1.00, 1.00] \\
                 & \textsc{rec-only} & 0.75 [0.71, 0.80] & 0.68 [0.63, 0.72] & 0.61 [0.55, 0.66] & 0.48 [0.42, 0.55] & 0.96 [0.94, 0.98] \\
                 & \textsc{kv-only}  & 0.09 [0.06, 0.11] & 0.21 [0.18, 0.25] & 0.08 [0.06, 0.12] & 0.08 [0.06, 0.10] & 0.82 [0.79, 0.85] \\
    \midrule
    Qwen3.5-27B$^\ddagger$ & \textsc{full}     & 0.00 [0.00, 0.00] & 0.00 [0.00, 0.00] & 0.00 [0.00, 0.00] & 0.00 [0.00, 0.00] & 1.00 [1.00, 1.00] \\
                 & \textsc{rec-only} & 0.52 [0.45, 0.59] & 0.20 [0.15, 0.26] & 0.50 [0.43, 0.57] & 0.52 [0.45, 0.59] & 0.59 [0.52, 0.66] \\
                 & \textsc{kv-only}  & 0.31 [0.25, 0.38] & 0.01 [0.00, 0.02] & 0.30 [0.23, 0.36] & 0.38 [0.31, 0.44] & 0.98 [0.95, 1.00] \\
    \midrule
    Falcon-H1-34B$^\ddagger$ & \textsc{full}     & 0.01 [0.00, 0.02] & 0.00 [0.00, 0.00] & 0.00 [0.00, 0.00] & 0.00 [0.00, 0.00] & 1.00 [1.00, 1.00] \\
                 & \textsc{rec-only} & 0.15 [0.10, 0.20] & 0.00 [0.00, 0.00] & 0.05 [0.02, 0.08] & 0.02 [0.01, 0.04] & 0.39 [0.32, 0.45] \\
                 & \textsc{kv-only}  & 0.17 [0.12, 0.22] & 0.02 [0.00, 0.04] & 0.12 [0.08, 0.17] & 0.12 [0.08, 0.17] & 0.68 [0.61, 0.74] \\
    \bottomrule
  \end{tabular}
  \caption{Memory-conjunction trap (Demo 4). Five rates per (model,
  mode): conjunction-absent FPR (the trap), neutral-absent FPR
  (yes-bias control), left-only and right-only FPR (only one of the
  two morphemes present in the list), and present TPR (compound itself
  in the list). Unmarked rows: $n{=}200$ per condition per seed,
  3 seeds, 95\% bootstrap CIs. $^\ddagger$Single-seed extension
  (seed 42 only); intervals are within-seed bootstraps.}
  \label{tab:app-demo-conj}
\end{table*}

\end{document}